\documentclass[twoside,11pt]{article}

\usepackage{jmlr2e}

\usepackage{color, colortbl}
\usepackage{longtable}
\usepackage{breakcites}
\usepackage{pdfpages}
\usepackage{rotating}
\usepackage{tikz}
\usetikzlibrary{arrows,calc,fit}
\definecolor{blue}{rgb}{0.5,0.5,0.5}

\usetikzlibrary{arrows.meta}

\usepackage[english]{babel}
\usepackage[utf8]{inputenc} % Spanish accent, and allows Bohmer15 german two dots to work! utf8x does not work!
\usepackage[T1]{fontenc}
\usepackage{stmaryrd} 

\usepackage[a4paper,top=3cm,bottom=2cm,left=3cm,right=3cm,marginparwidth=1.75cm]{geometry}

\usepackage{amsmath}
\usepackage{amsfonts}
\usepackage{graphicx}
\usepackage{subcaption}
\usepackage{mathtools}

\usepackage[colorinlistoftodos]{todonotes}
\usepackage{longtable} 
\usepackage{rotating} 
\usepackage{dsfont} % identity matrix symbol
\usepackage{xspace}

\definecolor{babyblueeyes}{rgb}{0.63, 0.79, 0.95}
\definecolor{blanchedalmond}{rgb}{1.0, 0.92, 0.8}
\definecolor{tangerine}{rgb}{0.95, 0.52, 0.0}

\newcommand{\gtcorr}{\textit{GTC}\xspace}
\newcommand{\gtcorrmean}{$GTC_{mean}$\xspace}

\ShortHeadings{State Representation Learning for Reinforcement Learning Toolbox}{A. Raffin et al. 2018}
\firstpageno{1}

\begin{document} 
\title{S-RL Toolbox: Environments, Datasets and Evaluation Metrics for State Representation Learning}

\author{\name Antonin Raffin \email antonin.raffin@ensta-paristech.fr
\AND
Ashley Hill \email ashley.hill@ensta-paristech.fr
\AND
Ren\'e Traor\'{e} \email rene.traore@ensta-paristech.fr
\AND
Timoth{\'{e}}e Lesort \email timothee.lesort@ensta-paristech.fr
\AND
Natalia D{\'{\i}}az-Rodr{\'{\i}}guez \email natalia.diaz@ensta-paristech.fr
\AND
\name David Filliat \email david.filliat@ensta-paristech.fr \\
\addr U2IS, ENSTA ParisTech / INRIA FLOWERS Team \url{http://flowers.inria.fr}\\
Palaiseau, France
}
\editor{%
}

%%%% USEFUL LINKS: 
% Links to graphs and figure editings in google drive https://drive.google.com/drive/folders/1DT25jDz3tR23RmA_ns17KRwLOj7ZRZ4E

% LINK TO ICRA PAPER FOR WHICH WE SHOULD HAVE SIMILAR STRUCTURE:
% https://www.overleaf.com/16783239vdwzkhfdkbzs#/64278669/
% blog post Antonin https://docs.google.com/document/d/1CeFmJjN61Irt2cLXScBQZJugcYh7AvN6RSfR4fspB4A/edit#
% review paper

\maketitle

\begin{abstract}%   <- trailing '%' for backward compatibility of .sty file
State representation learning aims at learning compact representations from raw observations in robotics and control applications. Approaches used for this objective are auto-encoders, learning forward models, inverse dynamics or learning using generic priors on the state characteristics. However, the diversity in applications and methods makes the field lack standard evaluation datasets, metrics and tasks. This paper provides a set of environments, data generators, robotic control tasks, metrics and tools to facilitate iterative state representation learning and evaluation in reinforcement learning settings. 
\end{abstract}


\section{Introduction}
Robotics control relies on compact and expressive representations of sensor data, as the task objectives are often expressed in much smaller dimensions than the sensor space dimension (e.g., the position of an object versus the size of an image). These representations are usually hand crafted by human experts, but deep-learning now makes it possible to avoid this feature engineering by using end-to-end learning (e.g., learning a policy from raw pixels). However, such approach is mostly possible in simulation because it requires a huge amount of training data --usually millions of samples-- that makes it impractical in the real world. To overcome this issue, State Representation Learning (SRL)  methods~\citep{Jonschkowski2018} can be used to create an intermediate representation that should contain only useful information to control a robot and thus simplify the policy learning task.

Many different SRL approaches have been proposed (see \citep{Lesort18} for a review), but comparing their performances is challenging. A common approach to evaluate learned representations is to compare performance in a Reinforcement Learning (RL) setting. However, because of the instability of RL algorithms and their cost, this should not be the only method used to assess learned states. Moreover, this approach gives no mean to interpret a state representation, making it difficult to understand which information is encoded in this representation.

While Reinforcement Learning has well established benchmarks, SRL has no metrics nor universal criterion to compare the different approaches. With that in mind, we propose a set of environments with an increasing difficulty, designed for comparing State Representation Learning (SRL) methods for robotic control. We also introduce qualitative and quantitative metrics along with visualization tools to facilitate the development and the comparison of SRL algorithms. The proposed framework allows fast iteration and eases research of new SRL methods by making it easy to produce statistically relevant results: the simulated environments run at 250 FPS on a 8-core machine that allows to train a RL agent on 1 Million steps in only 1h (or to generate 20k samples in less than 2 min)\footnote{Environments, code and data are available at \url{https://github.com/araffin/robotics-rl-srl}}.

In this paper, we first present quickly the reinforcement learning framework and the main state representation learning approaches that are implemented, before presenting the SRL Toolbox environments and datasets, the qualitative and quantitative evaluation methods, and a set of experiments illustrating the performances of the implemented approaches.

\section{Reinforcement Learning and State Representation Learning}

In this section, we introduce Reinforcement Learning (RL), as well as the State Representation Learning (SRL) approaches we integrated in our framework.

\subsection{Reinforcement Learning}

In RL, an agent must learn to select the best action to maximize a reward it will receive. More formally, in a given state $s_t$ ($t$ denotes the current time-step), an agent performs an action $a_t$ and receives a reward $r_t$. The learned behaviour, that should maximize the long-term discounted reward by mapping states to actions, is called a policy: $a_t = \pi(s_t)$.

In the most common settings, the state $s_t$ is either a low dimension representation given by an expert human, or corresponds to the raw observation (end-to-end learning). In order to differentiate these cases, we introduce the observation $o_t$ that correspond to the raw sensor data, and use the term \textit{state} only to refer to low dimensional representations that can be provided by humans, or can be learned using SRL.

In this paper, we work in the context of Markov Decision Processes (MDP), where the next state of the system only depends on the previous state and the taken action. As described above, the term \textit{observations} refers to raw sensor data (mostly images), but do not imply partial observability as assumed in Partially Observable Markov Decision Processes (POMDP).

\subsection{State Representation Learning}

\label{sec:SRL}
SRL~\citep{Lesort18} aims at learning compact representations from raw observations (e.g., learn a position $(x, y)$ directly from raw pixels) without explicit supervision. Most of the time, the goal is to use that representation to solve a task with RL. The idea is that a \textit{low-dimensional representation} should only keep the useful information and reduce the search space, thus contributing to address two main challenges of RL: sample inefficiency and instability. Moreover, a state representation learned for a particular task may be transferred to related tasks and therefore speed up learning in multiple task settings.

Using RL notations, SRL corresponds to learning a transformation\footnote{In practice, the learned transformation is a neural network} $\varphi$ from the observation space $\mathcal{O}$ to the state space $\mathcal{S}$.
Then, a policy $\pi$, that takes a state $s_t \in \mathcal{S}$ as input and outputs action $a_t$, is learned to solve the task:

\begin{equation}
o_t \xrightarrow[SRL]{\varphi} s_t \xrightarrow[RL]{\pi} a_t
\end{equation}

In the next sections, we present approaches of SRL that are implemented in our toolbox. Each method is not mutually exclusive and can be combined to create new models.

\subsubsection{Auto-encoders (AE, VAE)}
A first approach to learn a state representation is to compress the observation into a low dimensional state that is sufficient to reconstruct the observation. This approach does not take advantage of the robotic context because it ignores the possible actions, therefore it is often associated with different objectives (e.g. forward model), and provides a performance baseline.

We integrated Auto-Encoders \citep{Baldi12} and Variational Auto-encoders (VAE) \citep{Kingma13}, i.e., auto-encoders that enforce the latent variables to follow a given distribution.

\subsubsection{Robotic Priors}
\label{sec:priors}
To build a relevant representation of states, one can use prior knowledge about the dynamics or physics of the world. This knowledge can account for temporal continuity or causality principles that reflect the interactions of an agent with its environment \citep{Jonschkowski15, Jonschkowski17}. \textit{Robotic Priors} are defined as objective functions that constrain the state representation.

\subsubsection{Forward and Inverse Models}
The dynamics of the world can be integrated by learning a forward model that predicts state $s_{t+1}$ given state $s_t$ and action $a_t$. Constraints on the state representation can be added by constraining the forward model, for instance, by enforcing the system to follow linear dynamics~\citep{Watter15}.

Another approach is to learn an inverse model \citep{Shelhamer17,Pathak17}, which predicts the taken action $a_{t}$ given two successive states $s_t$ and $s_{t+1}$. This enforces the states to encode information about the dynamics, in order to recover the action needed for such transition.

\subsubsection{Combining Approaches}

Auto-encoders tend to reconstruct everything (that is salient enough in the observation), including static objects and irrelevant features for the task (distractors), whereas forward and inverse models focus on the dynamics, usually encoding the position of the controlled robot in the representation, but not a goal that is not controlled by the actions.

However, these approaches are not mutually exclusive and can be combined to create improved state representations. For instance, \citep{Pathak17} combine both inverse and forward models and \citep{Zhang2018} additionally integrate an auto-encoder.
In the same vein, \citet{Ha18} use a VAE with a recurrent forward model to learn a state representation.

\section{Datasets and Environments}
\label{sec:datasets}

In this section, we describe a set of environments with incremental difficulty, designed to assess SRL algorithms for robotic control. They all follow the interface defined by OpenAI Gym \citep{Brockman16}, which makes integration with RL algorithms easy.

\begin{figure}[h!]
\centering

  \begin{tabular}
      {ccc} \hline
      Mobile Navigation &
      Robotic Arm
      &
      Real Robot
      \\  \hline \\[0.0005pt]

\includegraphics[height=0.95in]{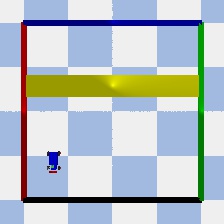} \includegraphics[height=0.95in]{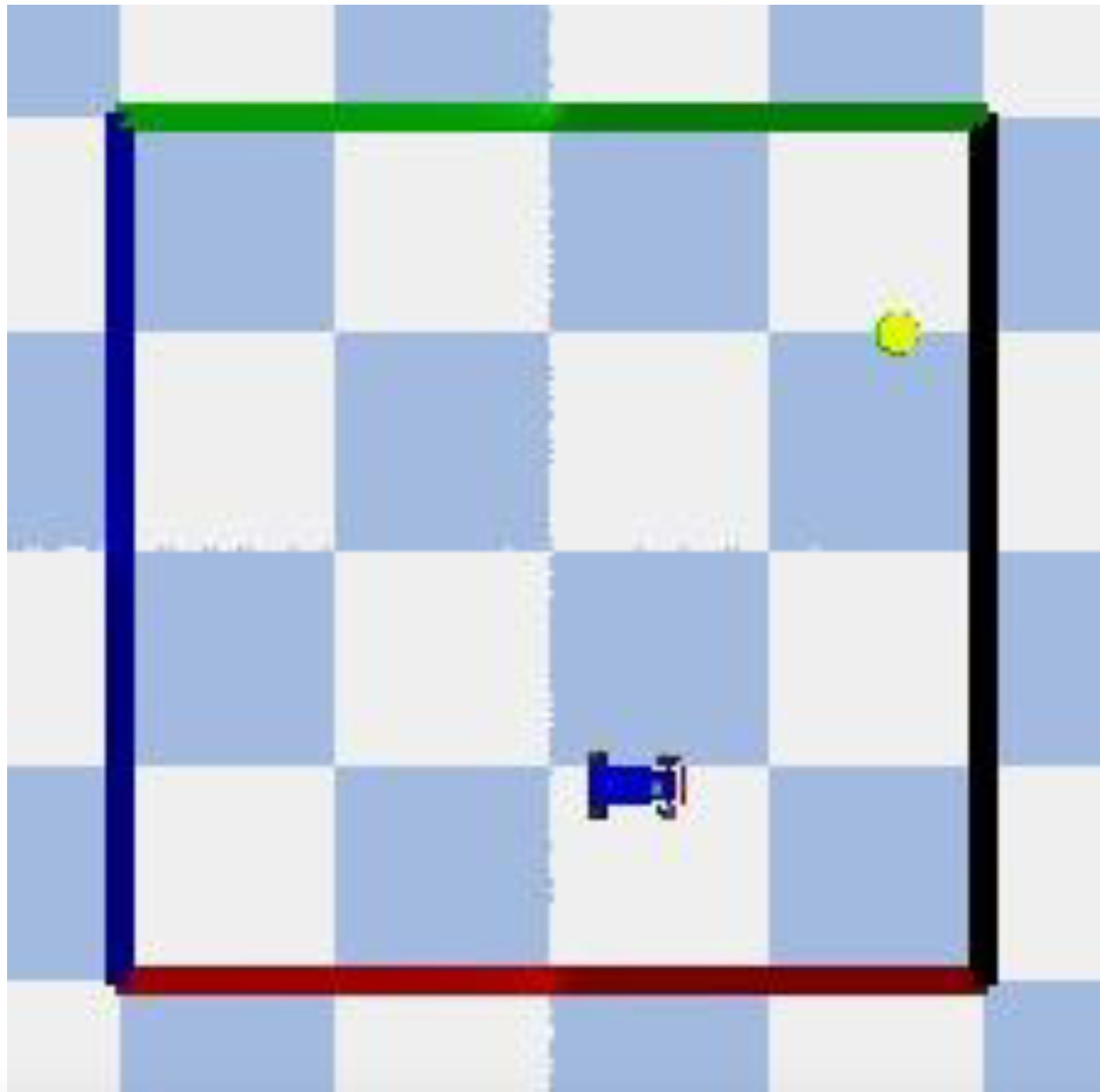} & \includegraphics[height=0.95in]{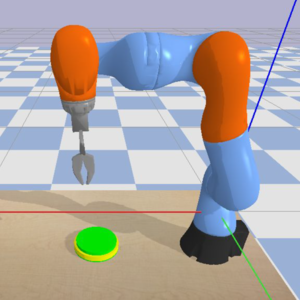}
& \includegraphics[height=0.95in]{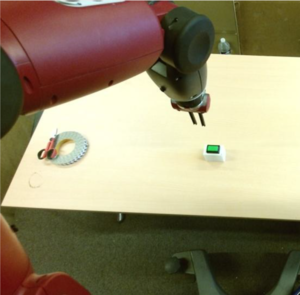} \\ \hline
  \end{tabular} 
  \caption{Environments and datasets for state representation learning.\label{fig:datasets}}
\end{figure}

\subsection{Environments Details}

The settings we propose (Fig.~\ref{fig:datasets}) are variations of two  environments: a 2D environment with a mobile robot and a 3D environment with a robotic arm. In all settings, there is a controlled robot and one or more targets (that can be static, randomly initialized or moving). Each environment can either have a continuous or discrete action space, and the reward can be sparse or shaped, allowing us to cover many different situations.

\textit{Static \& random target mobile navigation}: This setting  simulates a navigation task using a small car resembling the task of \citep{Jonschkowski-14-RSS}, with either a cylinder or a horizontal band on the ground as a goal, which can be fixed or moving from episode to episode. The car can move in four directions (\textit{forward, backward, left, right}) and will get a +1 reward when reaching the target, -1 when hitting walls, and 0 otherwise.

\textit{Static \& random target robotic arm}: This setting simulates a robotic arm (Kuka), fixed on a table, with the task of pushing a button that may move or not in between episodes. The arm can be controlled either in the $x$, $y$ and $z$ position using inverse kinematics, or directly controlling the joints. The robot will get +1 reward when the arm pushes the button, -1 when it hits the table (this will end the episode), and 0 otherwise. A variant adds distractors, i.e., moving objects on the table irrelevant for solving the task.

\textit{Simulated \& real robotic arm}: We used a real Baxter robot arm (Gazebo in simulation) to perform the same button pushing task as the previous task, with the same actions and the same rewards. The goal is to test the different methods in a real world setup.

A ground truth state is defined in each scenario: the absolute robot position in static scenarios and the relative position (w.r.t. the target) in moving goal scenarios. Note that apart from providing all described environments, we also provide the corresponding datasets
used in our evaluations: images are 224x224 pixels, navigation datasets use 4 discrete actions (\textit{right, left, forward, backward}); robot arms use one more (\textit{down}) action.

In section~\ref{sec:baselines-results}, we provide baselines results (Ground Truth, Auto-Encoder, Raw Pixels) for each environment.

\subsection{Motivation of the Goal-Based Robotics Tasks}

The environments proposed have several characteristics that make them suitable for research and benchmarking.

\textit{Designed for Robotics:} The proposed environments cover basic goal-based robotics tasks: navigation for a mobile robot and reaching a desired position for a robotic arm.

\textit{Designed for State Representation Learning:} The simplicity of the environments makes the extracted features easier to interpret (correlation can be computed between learned states and position of relevant objects). It is also clear what a good state representation should encode because of the small number of important elements: there is only the controllable robot and the target. 

\textit{Designed for Research:} The environments have incremental difficulty, the minimal number of variables for describing each environment (minimal state dimension for solving the task with RL) is increasing from 2 (mobile robot with static target) to 6 (robotic arm with random target).
Having simple environments of gradual difficulty is really important when developing new methods. The proposed environments are also easily customizable so that they cover all possibilities: reward can be sparse/dense, actions can be continuous/discrete. Finally, our benchmark is completely free and fast (it runs at 250 FPS on a 8-core machine with one GPU).

\section{Evaluation of Learned State Representations}
The most practical evaluation of SRL is assessing if the learned states can be used for solving the task in RL. However, algorithm development can benefit from other metrics and visually assessing the validity of the representation being learned, for faster iteration and interpretation of the state embedding space. We provide tools and metrics for this.

\subsection{Qualitative Evaluation}

\begin{figure}[h!]
\begin{center}

  \begin{tabular}
      {ccc} \hline  Real-time SRL & Interactive scatter  &  Latent visualization  \\   \hline \\[0.0005pt] 
      \begin{tikzpicture}[      
        every node/.style={anchor=south west,inner sep=0pt},
        x=1mm, y=1mm,
      ]   
     \node (fig1) at (0,0)
       {\includegraphics[width=1.65in]{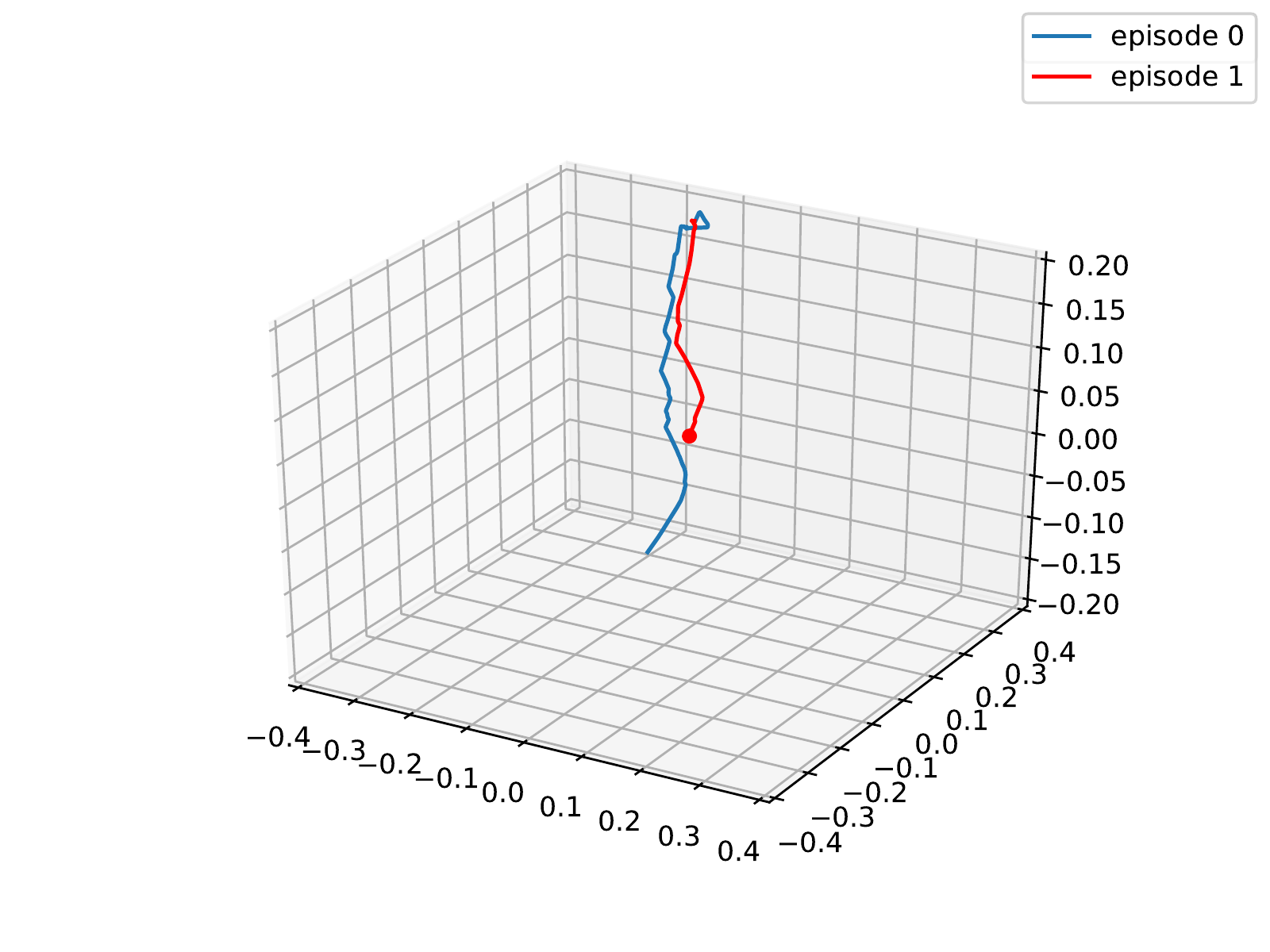}};
     \node (fig2) at (1,1)
       {\includegraphics[width=0.3in]{arxiv-images/kuka-simple-view}};  
	\end{tikzpicture} &
    \includegraphics[width=1.65in]{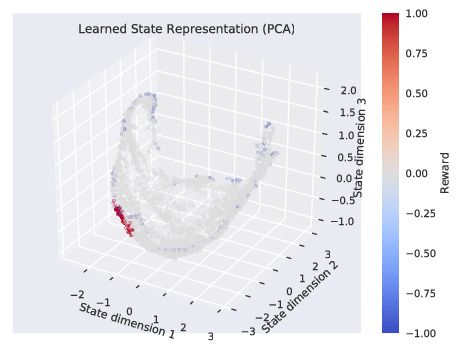} &
    \begin{tikzpicture}[      
        every node/.style={anchor=south west,inner sep=0pt},
        x=1mm, y=1mm,
      ]   
     \node (fig1) at (0,0)
       {\includegraphics[width=1.2in]{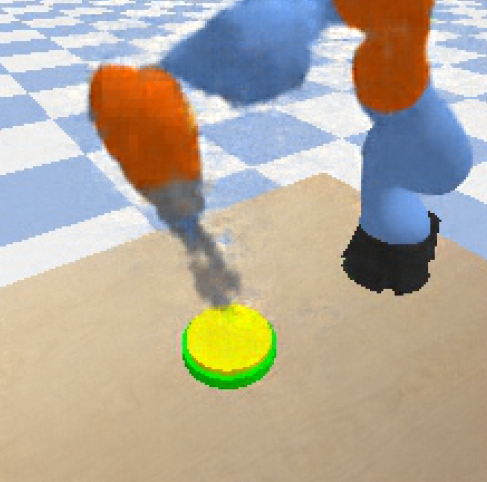}};
     \node (fig2) at (1,1)
       {\includegraphics[width=0.5in]{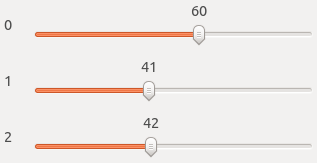}};  
	\end{tikzpicture}
    \\ \hline
  \end{tabular}
\end{center}
\caption{Visual tools for analysing SRL; Left: Live trajectory of the robot in the state space. Centre: 3D scatter plot of a state space; clicking on any point displays the corresponding observation. Right: reconstruction of the point in the state space defined by the sliders.
See complementary  material for videos. \label{fig:tools}}
\end{figure}

Qualitative evaluation in our case is the perceived utility of the state representation using visualization tools. The perceived utility depends on the task at hand. For example, the state representation of the robotic arm dataset is expected to have a continuous and correlated change with respect to the arm tip position. Three tools are proposed (Fig.~\ref{fig:tools}):

\textit{Real-time SRL}: This tool is used in conjunction with a graphical interface of the simulated environment. It shows the correspondence between observation and state by plotting current position of the observation in the state representation.

\textit{Interactive scatter}: A clickable plot of the state representation, with the reward defining the colour for each point. Here, we expect to visualize structure in the state representation.

\textit{Latent visualization}: It allows to navigate in the latent space by projecting the state to the observation space. This is achieved either by reconstructing the output (for AE and VAE), or by using a nearest neighbour approach for the models lacking a reconstruction.

To deal with state dimensions larger than three, PCA is used to visually assess the learned representations in a qualitative manner.

These tools allow to have better insights of the learned representation over different settings of state representation learning. This is especially useful in situations where the state dimension is greater than 3. This way, we are able to validate different configurations before running more exhaustive and time consuming methods.

\subsection{Metrics}
\label{sec:metrics}

\subsubsection{KNN-MSE}

We use an assessment of the representation's quality based on a Nearest-Neighbours approach (as in \citep{Sermanet17}).
While the nearest neighbour coherence can be assessed visually, KNN-MSE~\citep{Lesort17} derives a quantitative metric from this information.

For a given observation $o$, we find the nearest neighbours of its associated state $s$ in the learned state space, and project them in the ground truth state space. Then we compute the average distance to its neighbours in the latter space:

\begin{equation}\label{eq:knn_mse_crit}
\textrm{KNN-MSE}(s)=\frac{1}{k}\sum_{s' \in KNN(s,k) } || \tilde{s} - \tilde{s}' ||^2
\end{equation}
where $\textrm{KNN}(s,k)$ returns the $k$ nearest neighbours of $s$ (chosen with the Euclidean distance) in the learned state space $\mathcal{S}$, $\tilde{s}$ is the ground truth state associated to $s$, and $\tilde{s}'$ is the one associated to $s'$. A low KNN-MSE means that a neighbour in the ground truth is still a neighbour in the learned representation, and thus, local coherence is preserved.
 
\subsubsection{Correlation}
 
A Pearson $\rho$ correlation coefficient's matrix is computed  for each dimension pair ($s$, $\tilde{s}$), where $\tilde{s}$ is the ground truth (GT) state, $s$ the learned state, and $\mu_{s}$ and $\sigma_{s}$ are the mean and standard deviation, respectively, of state $s$:
\begin{equation}
\rho_{s, \tilde{s}} = \frac{ \mathbb{E}[ (s -\mu_{s}) * (\tilde{s} -\mu_{\tilde{s}})  ] } {\sigma_{s} * \sigma_{\tilde{s}}}
\end{equation}

\begin{figure}[h!]
  \begin{center}
  \includegraphics[width=2.8in]{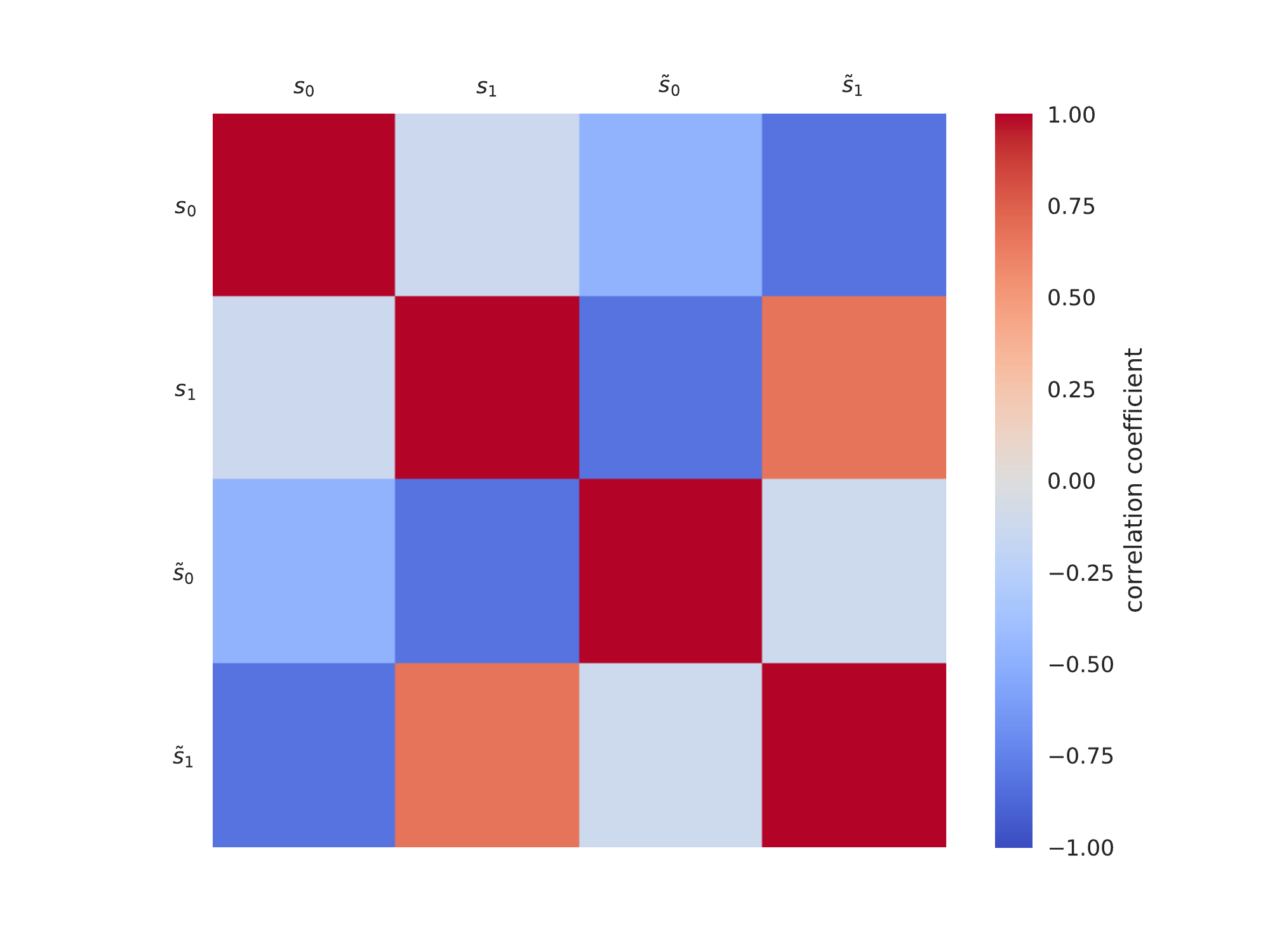}
  \caption{Correlation matrix for mobile robot navigation dataset (static target), between each dimension $s_i$ of predicted states $s$ and the ground truth $\tilde{s}_j$. We consider the ground truth to be the agent's real position. The states (dimension=2) are learned by combining a forward and an inverse model.}
  \label{fig:corr_matrix}
  \end{center}
 
\end{figure}

We can visualize the correlation matrix to quantitatively assess the ability of a model to encode relevant information in the states learned.
For instance, the correlation matrix in Figure \ref{fig:corr_matrix} shows degrees of correlation between the mobile robot position and the learned states.
The plot illustrates that for each dimension $i$ of the predicted states $s$, there is a correlation close to 1 (in absolute value) with at least one dimension $j$ of the agent's real position $\tilde{s}_j$. 
Therefore, this gives measurable evidences that the model was able to encode the position of the mobile robot.

This visualization tool is quite useful for low-dimensional spaces. However, for state spaces with a high number of dimensions, looking at the correlation matrix becomes impractical. Therefore, we introduce the following measure, named \gtcorr$\ $for Ground Truth Correlation, that allows to compare the models ability to encode relevant information:

\begin{equation}
GTC_{(i)} = \max\limits_j |\rho_{s, \tilde{s}} (i,j)| \in [0,1] 
\label{eq:c-vect}
\end{equation}

with $i \in \llbracket0, |\tilde{s}|\rrbracket$, $j \in \llbracket0,|s|\rrbracket$,
$\tilde{s} = [\tilde{s}_{1};...; \tilde{s}_{n}]$, and $\tilde{s}_{k}$ being the $k^{th}$ dimension of the ground truth state vector.
 
For instance, in the Mobile Robot environment with random target, the ground truth state is composed of the 2D robot position and 2D target position. That is to say, the ground truth states have a dimension of 4: $|\tilde{s}| = 4$.

The vector \gtcorr$\ $gives for each component $i$ of the ground truth states $\tilde{s}$, the maximum absolute correlation value between $\tilde{s}_i$ and any component of the predicted states $s$.
Therefore, \gtcorr$\ $measures the similarity per component, between the learned states $s$ and the ground truth states $\tilde{s}$.

We also introduce a metric, which is the mean of \gtcorr, that allows to compare learned states using one scalar value:

\begin{equation}
GTC_{mean} =\mathbb{E}[GTC]
\label{eq:c-mean}
\end{equation}

\subsection{Quantitative Evaluation With Reinforcement Learning}
\label{sec:quant}

Comparing the performance of RL algorithms, using the learned state representations, is the most relevant approach to evaluate the SRL methods. To do so, our framework integrates 8 algorithms (A2C, ACKTR, ACER, DQN, DDPG, PPO1, PPO2, TRPO) from Stable-Baselines~\citep{stable-baselines} (a fork of OpenAI baselines~\citep{Dhariwal17}), Augmented Random Search (ARS)~\citep{Mania18}, Covariance Matrix Adaptation Evolutionary Strategy (CMA-ES)~\citep{Hansen03} and Soft Actor Critic (SAC)~\citep{haarnoja2018}.

\section{Experiments}
We perform experiments on the proposed datasets using states learned with the approaches described in Section \ref{sec:SRL} along with ground truth (GT).
Here, we report results obtained with PPO.

%%%%%%%%%%%%%%%%%%%%
\begin{table}[h!]
  \begin{tabular}
      {lll} \hline %Env.  &  
      Ground Truth States & Learned States & RL Performance \\  \hline \\[0.0005pt]
\includegraphics[height=1.2in]{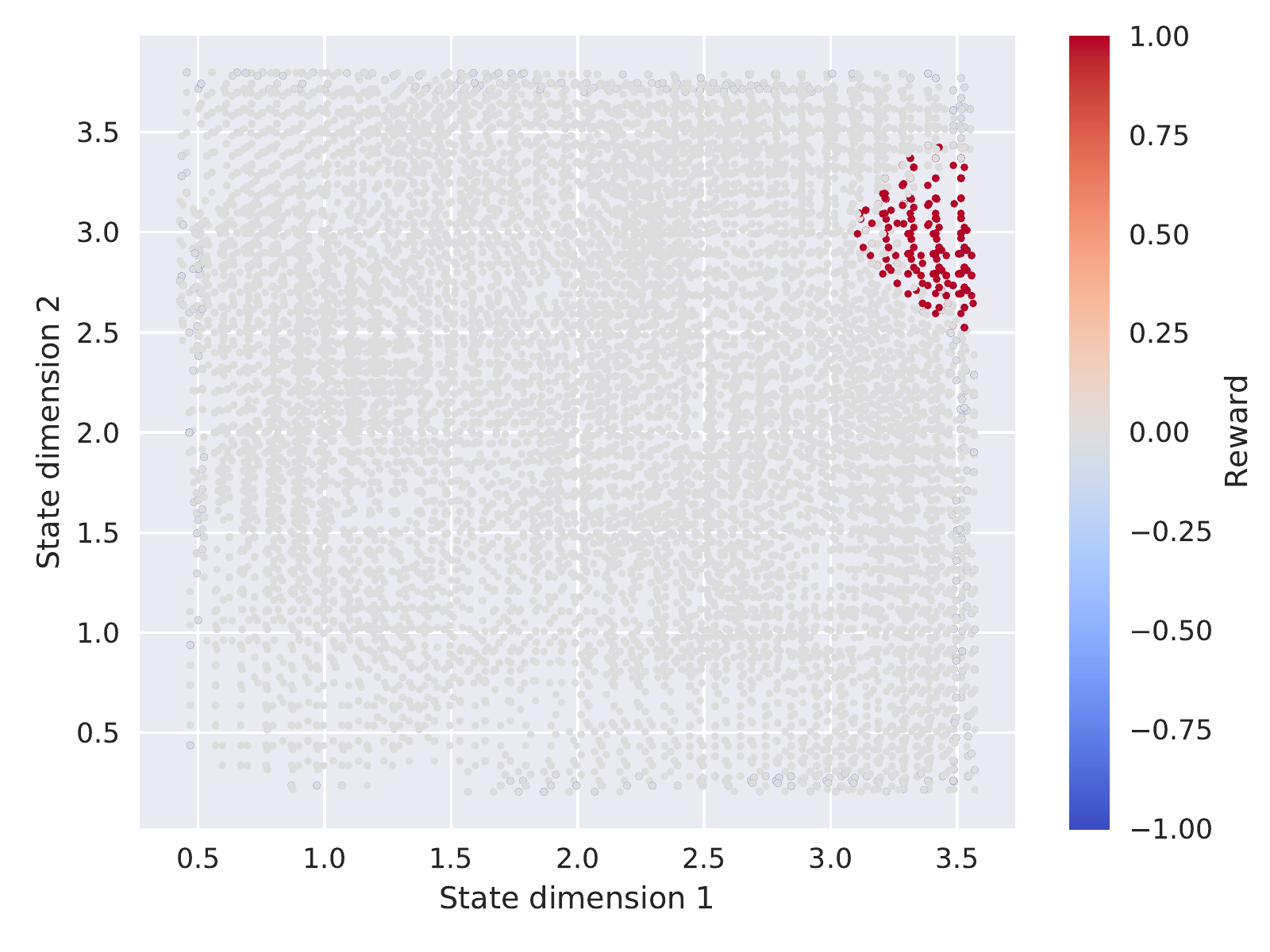} & \includegraphics[height=1.2in]{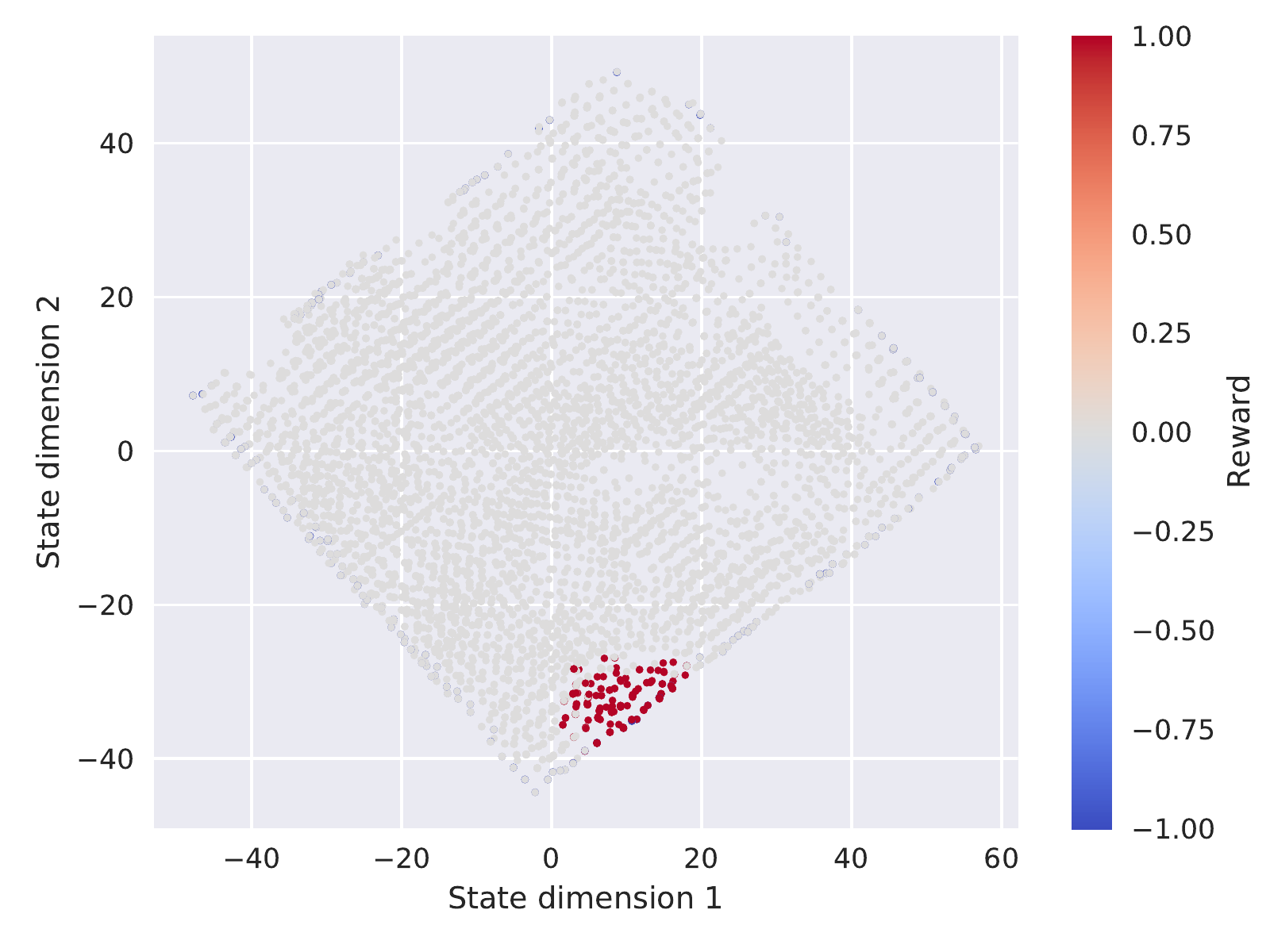} &
\includegraphics[height=1.5in]{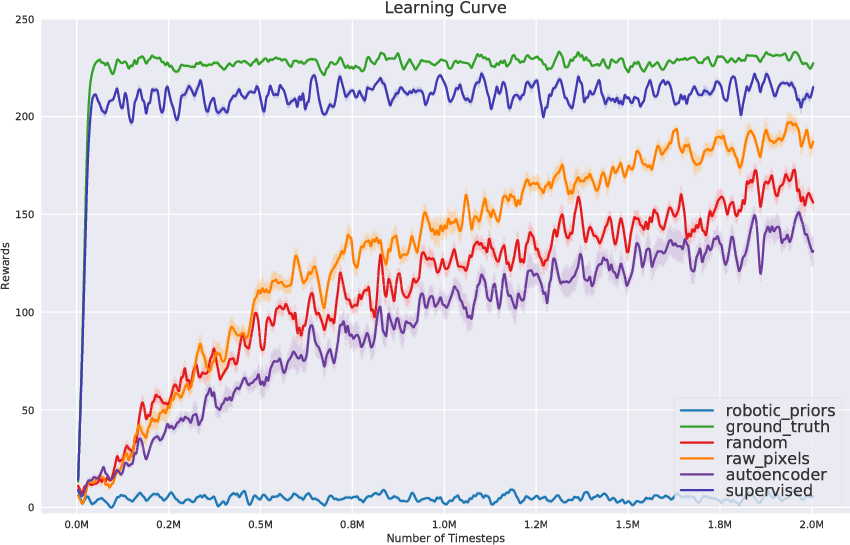} \\
      \hline
  \end{tabular}
  \centering 
\caption{Ground truth states (left), states learned (Inverse and Forward) (centre), and RL performance evaluation (PPO) (right) in the mobile robot environment. Colour denotes the reward, red for positive, blue for negative and grey for null reward (left and center). The full resolution plot of the RL performance evaluation can be found in appendix Fig. \ref{fig:ppo2-mobile-random-2M} (right) }
\label{tab:experiments}
\end{table}

Table \ref{tab:experiments} illustrate the qualitative evaluation of a state space learned by combining forward and inverse models on the mobile robot environment. It also shows the performance of PPO based on the states learned by several approaches.

%%%%%%%%%%%%%%%%%%%%%%%%%%%%%%
\begin{table} [h!]
 \centering 
\begin{tabular}{ lll|ll|ll}\hline 
\centering Dataset &   \multicolumn{2}{c}{\textit{Mobile-robot (1)}} &  \multicolumn{2}{c}{\textit{Robotic-arm  (3)}} &  \multicolumn{2}{c}{\textit{Robotic-arm-real (4)}}  \\ 
\centering Static/Random Target &  Static & Random & Static & Random & Static \\\hline 
\centering \textit{Ground Truth}  & 0.0099   & 0.0164   & 0.0025   & 0.0025  & 0.0105 \\ 
\centering \textit{AE} & 0.0168    & 0.7213     &  0.00336   &  0.0027  & 0.0179 \\
\centering \textit{VAE} & 0.0161     & 0.1295     & 0.0032    & 0.0027   & 0.0177 \\
\centering \textit{Robotic Priors} & 0.0200  & 0.0900 & 0.0029 & 0.0027 & 0.0213\\
\centering \textit{Forward} &  0.5111   &  1.1557    &  0.1425    & 0.2564   &  0.0796 \\
\centering \textit{Inverse} & 0.0191     &   0.7703   &  0.0182  & 0.2705   & 0.0521 \\
\centering \textit{Fwd+Inv} &  0.0164   &  0.7467 & 0.0176  & 0.2705   &  0.0368\\%
\hline 
\end{tabular}
\caption{KNN-MSE results for each SRL method. Ground Truth (robot absolute or relative position with respect to target object), auto-encoders, forward (fwd), inverse (inv) models and combinations.}
\label{tab:knn-mse}
\end{table}

Table \ref{tab:knn-mse} shows the KNN-MSE for the different SRL approaches on the implemented environments. % TODO

The table~\ref{tab:gt-correlation} gives the \gtcorrmean$ $metrics for several approaches and the associated RL performance using PPO. It shows that \gtcorrmean$ $is a good indicator for the performance than can be obtained in RL: a good disentanglement with a good correlation with the ground truth state (i.e. a high \gtcorrmean) lead to a higher mean reward in RL.

During the development of \textit{S-RL Toolbox}, we learned some useful insights on SRL. We observed that auto-encoders do not reconstruct objects if they are too small. This is an issue if the object is relevant for the task. An inverse model is usually sufficient to learn a coherent state representation (in our case, the position of the controllable object). Continuity in the state space is important to perform well in RL. Using real-time SRL tools, we note discontinuities in the state space learned by AE, even if reconstruction error is low, and that hinders RL.

%%%%%%%%%%%%%%%%%%%%%%%%%%%%%%
%% TABLE: GT Correlation 
% TODO: Add autoencoder + reward
\begin{table} [h!]
 \centering 
\begin{tabular}{ l|llll|l|l}\hline 
GroundTruthCorrelation & \textit{$x_{robot}$} & \textit{$y_{robot}$} & \textit{$x_{target}$}  & \textit{$y_{target}$} & \textit{Mean} & \textit{RL} \\\hline  
Robotic Priors & 0.2 & 0.2 & 0.41 & 0.66 & 0.37 & 5.4 $\pm$ 3.1 \\
Random & 0.68 & 0.65 & 0.34 & 0.31 & 0.50 & 163.4 $\pm$ 10.0 \\
Supervised & 0.69 & 0.73 & 0.70 & 0.72 & 0.71 & 213.3 $\pm$ 6.0 \\
Auto-Encoder & 0.52 & 0.51 & 0.24 & 0.23 & 0.38 & 138.5 $\pm$ 12.3 \\
GT  &  1 &  1 & 1 & 1 & 1 & 229.7 +- 2.7 \\
\hline 
\end{tabular}
\caption{\gtcorr and \gtcorrmean, and mean reward performance in RL (using PPO) per episode after 2 millions steps, with standard error for each SRL method in mobile robot with random target environment.}
\label{tab:gt-correlation}
\end{table}
%%%%%%%%%%%%%%%%
%%%%%%%%%%%%%%%

\section{Related Work}

In the RL literature, several classic benchmarks are used to compare algorithms. For discrete actions, Atari Games from OpenAI Gym suite~\citep{Brockman16} are often adopted, whereas for continuous actions, MuJoCo locomotion tasks \citep{Todorov2012} is favoured. OpenAI recently open-sourced robotics goal-based tasks \citep{Plappert18}. These environments are close to what we propose; however, they use a non-free physics engine, observations are not images and customization is not easy. Also, our tools were designed with SRL in mind and offer a gradual difficulty.
%Talk about the advantage of having integrated tools

Regarding the SRL literature, very diverse environments are used, without common metrics or visualizations. The SRL methods are usually only compared with learning from raw observations. \cite{Jonschkowski15} assess the quality of learned representations in a slot car scenario (similar to \cite{Lange2012}) and a mobile navigation task (following \cite{Sprague2009, Boots2011}). More recently, \cite{Ha18} present their results on Vizdoom and RaceCar environments. They show interesting insights of what was learned by navigating in the latent space and projecting states back to the pixel space (similar to Fig.~\ref{fig:tools}). \cite{Zhang2018} use a subset of MuJoCo tasks, with joints as input, and a binary maze environment. They perform an insightful ablation study of their SRL model.

Our contribution is two-fold: we provide a framework integrated with RL algorithms, environments and tools for SRL, along with implementations of the main SRL methods.

\section{Discussion and Conclusions}
This paper presents a set of environments on which to perform SRL benchmarks of incremental difficulty to solve tasks in RL, specifically, in robotics control. Our proposed toolbox facilitates fast iterations, interpretability and reproducibility, with a set of qualitative and quantitative metrics and  interactive visualization tools. We believe such a framework is needed to have fair comparisons, focused on robotics control, among SRL methods.

\acks{This work is supported by the DREAM project\footnote{\url{http://www.robotsthatdream.eu}} through the European Union Horizon 2020 FET research and innovation program under grant agreement No 640891.}


\newpage

\appendix
\section{Implementation Details}

\subsection{Datasets Details}
In this section, we provide the parameters used to generate datasets, for the results presented in Table~\ref{tab:knn-mse}. The simulated datasets were created with a random policy, using \textit{PyBullet} \citep{Coumans18} (generating up to 30k samples  per minute on a 8-core machine~\footnote{CPU: Intel Core i7-7700K GPU:Nvidia GeForce GTX 1080 Ti}). The real Baxter dataset was recorded using ROS. 

We used 10k samples of each dataset to learn a state representation. Reward is sparse (see Section~\ref{sec:datasets}  for details) and actions are discrete (encoded as integers) for all datasets ~\footnote{Datasets can also be generated with continuous actions and dense (shaped) reward}.

Each target object can be fixed or randomly positioned between episodes; what we call "absolute" or "relative" dataset, refers to the fact of learning a relative or absolute position of the robot with respect to the target object. The images dimension in every dataset is 224x224 pixels.

\begin{table}[h!]
 \centering 
\caption{Dataset details. Distractors can both be static or moving}
\label{tab:environment-details-distractors}
\begin{tabular}
{lllll} \hline Dataset & Reward & Actions & Distractors  \\\hline
      
Mobile-robot & Sparse &  4, Discrete ($\Delta$(X,Y) pos.) & No \\
      
Robotic-arm & Sparse & 5, Discrete ($\Delta$(X,Y,Z) pos.) & No \\

Robotic-arm-real & Sparse & 5, Discrete ($\Delta$(X,Y,Z) pos.) & Yes \\  
      \hline
\end{tabular}
\end{table}

\subsection{Baselines Results}
\label{sec:baselines-results}
In this section, we provide baselines results for each environment.

%%%%%%%%%%%%%%%%%%%%%%%%%%%%%%
%% TABLE: RL performance Mobile Robot Line Target
\begin{table} [htbp!]
\centering 
\begin{tabular}{ l|ll}\hline 
\textbf{Budget} (in timesteps) & \textit{1 Million} & \textit{2 Million}\\\hline
Ground Truth  & 198.0 $\pm$ 16.1 & 211.6 $\pm$ 14.0  \\
\hline
Raw Pixels & 177.9 $\pm$ 15.6 & 215.7 $\pm$ 9.6 \\
Auto-Encoder & 159.8 $\pm$ 16.1 & 188.8 $\pm$ 13.5  \\
\hline
\end{tabular}
\caption{Mean reward performance in RL (using PPO) per episode (average on 100 episodes) for different budgets, with standard error in Navigation 1D target environment.}
\label{tab:rl-perf-mobile-1d}
\end{table}

%%%%%%%%%%%%%%%%%%%%%%%%%%%%%%
%% TABLE: RL performance Mobile robot
\begin{table} [htbp!]
 \centering 
\begin{tabular}{ l|llll}\hline 
\textbf{Budget} (in timesteps) & \textit{1 Million} & \textit{2 Million} & \textit{3 Million}  & \textit{5 Million}\\\hline
Ground Truth  & 227.8 $\pm$ 2.8 & 229.7 $\pm$ 2.7 & 231.5 $\pm$ 1.9 & 234.4 $\pm$ 1.3 \\
\hline
Raw Pixels & 136.3 $\pm$ 11.5 & 188.2 $\pm$ 9.4 & 214.0 $\pm$ 5.9 & 231.5 $\pm$ 3.1 \\
Auto-Encoder & 97.0 $\pm$ 12.3 & 138.5 $\pm$ 12.3 & 167.7 $\pm$ 11.1 & 192.6 $\pm$ 8.9 \\
\hline
\end{tabular}
\caption{Mean reward performance in RL (using PPO) per episode (average on 100 episodes) for different budgets, with standard error in Navigation 2D random target environment.}
\label{tab:rl-perf-mobile-2d}
\end{table}

%%%%%%%%%%%%%%%%%%%%%%%%%%%%%%
%% TABLE: RL performance kuka target
\begin{table}[htbp!]
 \centering 
\begin{tabular}{l|llll}\hline 
\textbf{Budget} (in timesteps) & \textit{1 Million} & \textit{2 Million} & \textit{3 Million}  & \textit{5 Million}\\\hline
Ground Truth  & 4.1 $\pm$ 0.5 & 4.1 $\pm$ 0.6 & 4.1 $\pm$ 0.6 & 4.2 $\pm$ 0.5 \\
\hline
Raw Pixels & 0.6 $\pm$ 0.3 & 0.8 $\pm$ 0.3 & 1.2 $\pm$ 0.3 & 2.6 $\pm$ 0.3 \\
Auto-Encoder & 0.92 $\pm$ 0.3 & 1.6 $\pm$ 0.3 & 2.2 $\pm$ 0.3 & 3.4 $\pm$ 0.3 \\
\hline
\end{tabular}
\caption{Mean reward performance in RL (using PPO) per episode (average on 100 episodes) for different budgets, with standard error in robotic arm with random target environment.}
\label{tab:rl-perf-kuka-target}
\end{table}

%%%%%%%%%%%%%%%%%%%%%%%%%%%%%%
%% TABLE: RL performance kuka moving
\begin{table}[htbp!]
\centering 
\begin{tabular}{l|llll}\hline 
\textbf{Budget} (in timesteps) & \textit{1 Million} & \textit{2 Million} & \textit{3 Million}  & \textit{5 Million}\\\hline
Ground Truth  & 4.3 $\pm$ 0.3 & 4.4 $\pm$ 0.2 & 4.4 $\pm$ 0.2 & 4.6 $\pm$ 0.2 \\
\hline
Raw Pixels & 0.8 $\pm$ 0.3 & 1.0 $\pm$ 0.3 & 1.2 $\pm$ 0.3 & 2.0 $\pm$ 0.3 \\
Auto-Encoder & 1.17 $\pm$ 0.3 & 1.5 $\pm$ 0.3 & 1.9 $\pm$ 0.4 & 3.0 $\pm$ 0.4 \\
\hline
\end{tabular}
\caption{Mean reward performance in RL (using PPO) per episode (average on 100 episodes) for different budgets, with standard error in robotic arm with moving target environment.}
\label{tab:rl-perf-kuka-moving}
\end{table}

\clearpage

\subsection{Reinforcement Learning Experiments}

\subsubsection{Reinforcement Learning}
We use the implementations from Stable-Baselines~\citep{stable-baselines} for the RL experiments (except for CMA-ES, ARS and SAC which we implemented ourselves). Where we used the default hyperparameters set by OpenAI Baselines. With 10 random seeds in order to have quantitative results.

As for the network of the policies, the same architecture is used in all different methods. For the SRL and ground truth approaches, it is a 2-layers MLP, whereas for learning from raw pixels, it is the CNN from \citep{Mnih15} implemented in OpenAI baselines.

All CNN policies normalize the input image by dividing it by 255. Observations are not stacked. When learning from SRL, the states are normalized using a running mean/std average.
Reinforcement learning metrics reported are the average returned rewards over 5 policies, independently trained using the same RL algorithm with a different seed.

\subsubsection{State Learning Representation}
For the SRL methods, we used an altered form of ResNet (Table \ref{tab:arch}), which is more compact for our needs. However, for forward and inverse models, we used a linear model in addition to the CNN for learning a state representation. We are using the Adam~\citep{Kingma14} optimizer in all our models, with learning rates that vary from $10^{-2}$ to $10^{-3}$. The batch size of forward and inverse is 128, 256 for robotic priors, and 32 for the other models.

\begin{table}[h!] 
\caption{CNN architecture for SRL methods. The first layers are inspired by ResNet architecture~\citep{He15}. }
\label{tab:arch}
\begin{center}
\begin{small}
\begin{tabular}{ll}
\hline
Layer & Architecture \\ \hline
1 & Conv2d(3, 64, kernel\_size=7, stride=2, padding=3, bias=False) + BN + ReLU \\
2 & MaxPool2d(kernel\_size=3, stride=2, padding=1) \\
3 & Conv2d(64, 64, kernel\_size=3, stride=1, padding=1, bias=False) + BN + ReLU \\
4 & MaxPool2d(kernel\_size=3, stride=2) \\
5 & Conv2d(64, 64, kernel\_size=3, stride=2, padding=1, bias=False) + BN + ReLU \\
6 & MaxPool2d(kernel\_size=3, stride=2) \\
7 & Linear(6 * 6 * 64, state\_dim) \\ \hline
\end{tabular}
\end{small}
\end{center}
\end{table}

%We evaluate different SRL approaches using DQN~\citep{Mnih15} and variants, ACER~\citep{Wang16}, A2C~\citep{Wang16learning}, CMA-ES~\citep{Hansen03} and PPO~\footnote{in our plots, we used PPO2, an optimized version for GPU from OpenAI Baselines}.

\subsubsection{Results}

\begin{figure}[h!]
 \centering 
 \includegraphics[scale=0.5]{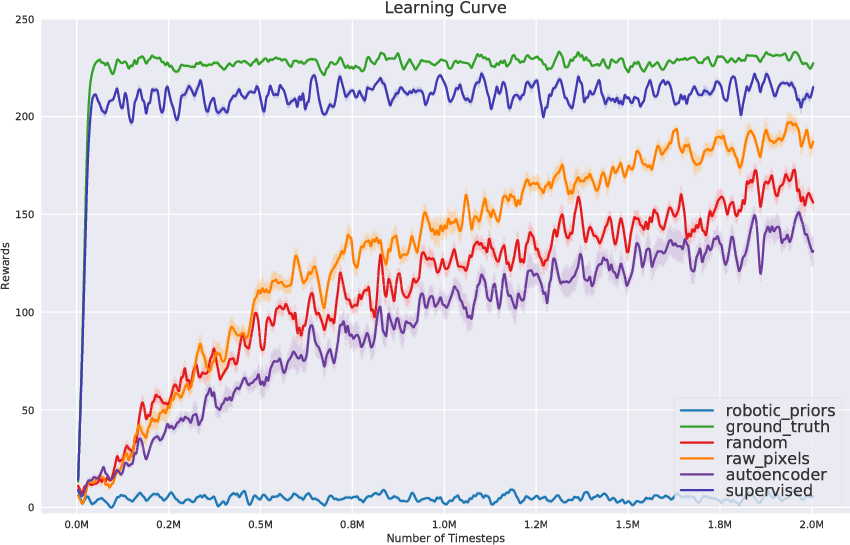}
 \caption{Performance (mean and standard error for 10 runs) for PPO algorithm for different state representations learned in mobile-robot-navigation (random target) environment}
 \label{fig:ppo2-mobile-random-2M}
\end{figure}

Fig. \ref{fig:ppo2-mobile-random-2M} shows the average reward during learning of the PPO algorithm.

\begin{figure}[h!]
 \centering 
 \includegraphics[scale=0.5]{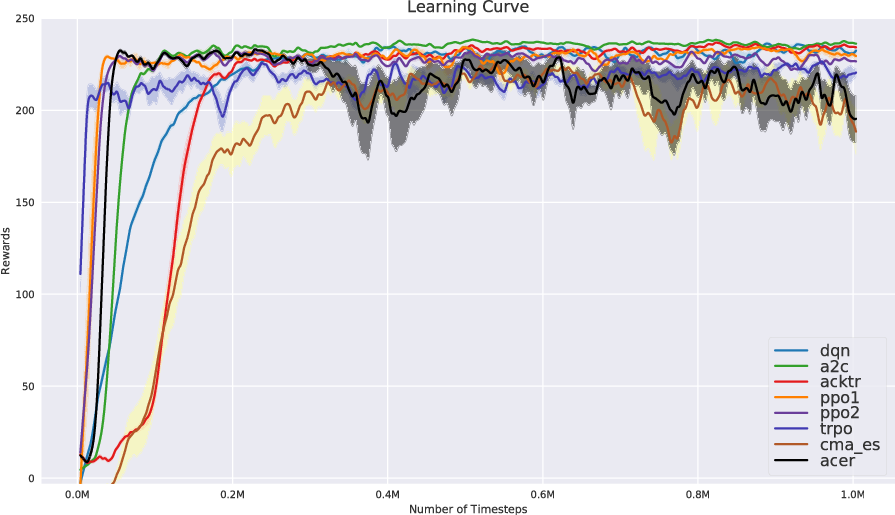}
 \caption{Performance (mean and standard error) on RL algorithms using ground truth states with  mobile robot (random target) environment} 
 \label{fig:rl-benchmark}
\end{figure}

We observed during the development of the \emph{S-RL Toolbox}, that PPO was one of the best RL algorithm for SRL benchmarking. It allows us to obtain good performance and was consistent without needing to change any hyperparameters. Figure \ref{fig:rl-benchmark} illustrates this, in addition to the variability of the RL algorithms.

\section{Supplementary material}
The visualization and interactive state space exploration tools are demonstrated in the following videos:

\begin{itemize}
\item S-RL Toolbox Showcase: \url{https://youtu.be/qNsHMkIsqJc}
\item S-RL Toolbox Environments: \url{https://tinyurl.com/y973vhfy}
\item Kuka robot arm: RL running PPO (SRL trained with VAE): \url{https://tinyurl.com/yarpbs2c}

\item Kuka robot arm environment: State Representation Learning Benchmark running PPO on ground truth states: \url{https://tinyurl.com/yd549l7o}
\end{itemize}

\vskip 0.2in
\bibliography{SRL4RL}

\end{document}